\documentclass[runningheads]{llncs}

\usepackage{eccv}
\usepackage{eccvabbrv}
\usepackage[accsupp]{axessibility} 
\usepackage{epsfig}
\usepackage{amsmath}
\usepackage{amssymb}
\usepackage{microtype}
\usepackage{enumitem}
\usepackage{makecell}
\usepackage{caption}
\usepackage{pifont}
\usepackage{url}
\usepackage[table]{xcolor}
\usepackage{graphicx,wrapfig,lipsum}
\usepackage{threeparttable}
\usepackage{tabularx}
\usepackage{booktabs}
\usepackage{multirow}
\usepackage{array}
\usepackage{colortbl}
\usepackage[breaklinks,colorlinks,citecolor=eccvblue]{hyperref}
\usepackage{orcidlink}

\usepackage{tcolorbox}
\newcommand{\PredSty}[1]
{\textnormal{\ttfamily\color{magenta!80!black}#1}\unskip}

\setitemize{noitemsep,topsep=0pt,parsep=0pt,partopsep=0pt}

\newcommand{\cmark}{\ding{51}}
\newcommand{\xmark}{\ding{55}}

\definecolor{bestorange}{HTML}{FFDDC1}
\definecolor{secondbestorange}{HTML}{FFF5EE} 

\newcommand{\best}[1]{\cellcolor{bestorange}\textbf{#1}}
\newcommand{\secondbest}[1]{\cellcolor{secondbestorange}#1}

\def\VspaceS{\vspace{-0.10cm}}
\def\VspaceL{\vspace{-0.10cm}}

\begin{document}

\title{3D-Aware VLMs with Implicit and Explicit Geometries}

\author{
Wenhao Li\inst{1},
Xueying Jiang\inst{1},
Quanhao Qian\inst{2,3},
Deli Zhao\inst{2,3}, \\
Ran Xu\inst{2,3}\textsuperscript{\ding{41}}, 
Shijian Lu\inst{1}\textsuperscript{\ding{41}},
Gongjie Zhang\inst{4}\textsuperscript{\ding{41}}
\\
\institute{ 
\textsuperscript{1}{\normalsize Nanyang Technological University} \\
\textsuperscript{2}{\normalsize DAMO Academy, Alibaba Group} \quad
\textsuperscript{3}{\normalsize HuPan Lab} \quad
\textsuperscript{4}{\normalsize Alibaba Group} \\
}}

{\renewcommand{\thefootnote}{\ding{41}}
\footnotetext{Corresponding Authors.}}

\authorrunning{Li et al.}

\maketitle

\begin{abstract}
Despite rapid progress, most existing vision-language models (VLMs) built from 2D visual inputs often struggle when handling various 3D tasks that require fine-grained spatial understanding and reasoning. To bridge this gap, we present VLM-IE3D, a unified framework that enhances the 3D spatial awareness of VLMs by equipping them with both implicit and explicit 3D geometries learned from RGB videos. Our VLM-IE3D introduces Implicit Geometry Tokens (IGTs) that capture high-level geometric priors from input videos, as well as complementary Explicit Geometry Tokens (EGTs) that encode detailed geometric structures from reconstructed 3D attributes. On top of that, VLM-IE3D comes with a 3D-aware adapter that effectively fuses the two types of geometric representations with 2D visual cues. This RGB-only design injects strong 3D inductive biases for fine-grained spatial understanding and reasoning without requiring any additional 3D inputs. Extensive experiments show that VLM-IE3D achieves superior performance consistently across various 3D tasks including 3D video detection, 3D visual grounding, 3D dense captioning, and spatial reasoning. Code and models are available at \url{https://github.com/Vegetebird/VLM-IE3D}. 
\end{abstract}

\section{Introduction}

Vision-language models (VLMs) have demonstrated great potential in various spatial understanding and reasoning tasks such as embodied navigation \cite{zhou2024navgpt,zhang2024vision}, vision-language-action (VLA) \cite{kim2024openvla,ma2024survey}, and 3D scene understanding \cite{jia2024sceneverse,zhang2024vision11}. 
Recently, one promising initiative \cite{vgllm,fan2025vlm,wu2025spatial,huang2025mllms} attempts to learn 3D representations directly from 2D visual observations instead of explicit 3D data (\textit{e.g.}, point clouds), mitigating the dependency on specialized 3D sensors and enhancing VLM applicability with more accessible 2D RGB data in real-world scenarios. 
One prevalent approach endows VLMs with 3D spatial awareness by introducing a 3D geometry encoder \cite{vggt,anysplat} to learn implicit 3D representations from video sequences. 
However, the learned implicit 3D representations often capture coarse and global spatial layouts of scenes, which are insufficient for handling various 3D understanding tasks that require fine-grained scene structures and geometries for precise spatial localization and reasoning. 

This limitation mirrors insights from cognitive science \cite{johnson1980mental,johnson1983mental}, where humans form abstract ``3D cognitive maps'' for coarse scene awareness, often without geometric details. 
Similarly, the implicit representations from 3D geometry encoders in current VLMs provide high-level spatial priors. 
However, their compressed and latent nature makes it difficult for a language model to interpret quantitative geometric properties, even if such information is latently encoded. This leaves fine-grained geometric information inaccessible for precise reasoning tasks in existing VLMs. 

\begin{figure*}[t]
\centering
\includegraphics[width=1.00\linewidth]{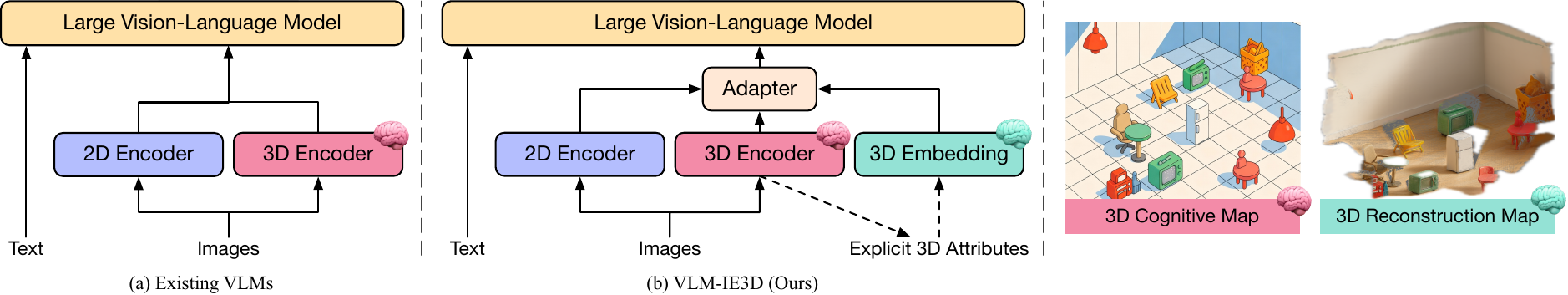}
\caption
{
\textbf{(a)} Existing VLMs acquire 3D awareness by learning solely implicit geometric representations that provide a coarse ``3D cognitive map''. \textbf{(b)} The proposed VLM-IE3D introduces explicit geometric representations from reconstructed 3D attributes, offering a ``3D reconstruction map'' that is equipped with fine-grained local geometries that are supplementary to the implicit representations. 
These two complementary representations are then fused with 2D visual cues, enhancing the 3D awareness of VLMs with both coarse global layouts and fine-grained local geometric details. 
}
\label{fig:teaser}
\VspaceL
\end{figure*}

In contrast, robotic systems typically rely on explicit 3D representations (\textit{e.g.}, depth maps and point clouds) that form structured ``3D reconstruction maps'', making quantitative geometric data directly accessible for fine-grained spatial tasks such as navigation and localization. 
This structured and interpretable form of geometric representations is crucial for precise spatial understanding, yet remains largely unexploited in RGB-only VLM frameworks.  
Our work is motivated by bridging this gap: equipping VLMs with both the abstract spatial awareness of ``3D cognitive maps'' (global priors) and the fine-grained structured understanding of ``3D reconstruction maps'' (fine-grained structures) (see Fig.~\ref{fig:teaser}). 

To address this gap, we propose VLM-IE3D, a unified framework that equips VLMs with both implicit and explicit 3D geometric representations extracted from RGB videos. 
Unlike prior works \cite{vgllm,fan2025vlm,wu2025spatial,huang2025mllms} that solely rely on implicit 3D representations, VLM-IE3D introduces two complementary geometry-aware representations to intrinsically inject strong 3D inductive biases into VLMs:
\textbf{(i)} Implicit Geometry Tokens (IGTs), generated by a 3D geometry encoder, capturing coarse and high-level spatial information for global scene understanding;
\textbf{(ii)} Explicit Geometry Tokens (EGTs), encoded by a lightweight embedding module from reconstructed explicit 3D representations (\textit{e.g.}, depth maps and point clouds), preserving detailed structural information for fine-grained spatial understanding. 
These two types of tokens are naturally complementary: IGTs provide high-level and coarse global geometric priors while EGTs capture specific and fine-grained local geometric structures. 
Furthermore, we develop a 3D-aware adapter to effectively integrate these implicit and explicit 3D representations with 2D visual cues, producing unified 3D-aware visual embeddings that enable VLMs to perform comprehensive spatial understanding and reasoning at both holistic and fine-grained levels without requiring additional 3D inputs. 

We conduct extensive experiments across diverse 3D understanding and reasoning tasks: 3D video object detection, 3D visual grounding, 3D dense captioning, and spatial reasoning. 
Results demonstrate that VLM-IE3D achieves state-of-the-art performance among RGB-only methods, validating the effectiveness of combining implicit and explicit geometric representations. 

Our main contributions are summarized in three aspects. 
\textit{First}, we present VLM-IE3D, a unified framework that leverages both implicit and explicit geometric representations to intrinsically inject strong 3D inductive biases into VLMs from purely RGB video sequences. 
\textit{Second}, we introduce two complementary geometry-aware representations: IGTs encode high-level 3D priors and EGTs provide detailed 3D structures. We also design a 3D-aware adapter to effectively integrate these representations with 2D visual cues. 
\textit{Third}, extensive experiments demonstrate that VLM-IE3D achieves competitive performance on multiple 3D scene understanding and spatial reasoning tasks without additional 3D inputs. 

\section{Related Work}

\noindent \textbf{Vision-Language Models.}
Vision-Language Models (VLMs) \cite{alayrac2022flamingo,liu2023llava,liu2024improved,peng2023kosmos} have achieved remarkable success in integrating vision and language. 
Early works such as CLIP \cite{radford2021learning} employ contrastive learning to align image and text representations, while subsequent models like BLIP \cite{li2022blip} enhance CLIP with larger datasets and improved training strategies. 
Recent developments \cite{lin2023videollava,li2023videochat,zhang2023video} have extended VLMs from static images to dynamic video.
For example, Qwen2.5-VL \cite{Qwen2.5-VL} strengthens temporal reasoning through dynamic resolution mechanisms and absolute time encoding. 
Despite being pre-trained on large-scale image–text corpora, these models are not explicitly optimized to capture 3D geometric information, thereby limiting their 3D capabilities. 

\noindent \textbf{3D Vision-Language Models with 3D Inputs.}
Recent progress in VLMs has extended their capabilities from 2D to 3D scene understanding by leveraging explicit 3D data as inputs (\textit{e.g.}, depth maps and point clouds) \cite{grounded-3dllm,fu2025scene,qi2025gpt4scene,wang2025ross3d}. 
For example, 3D-LLM \cite{hong20233d} injects 3D scene representations, obtained by aggregating multi-view 2D features, into a large language model for 3D reasoning. 
PointLLM \cite{xu2024pointllm} combines a point encoder with a large language model to fuse geometric, appearance, and linguistic information for point cloud understanding. 
Video-3D LLM \cite{video3dllm} backprojects each pixel in depth maps into 3D coordinates and injects these spatial cues into video features. 
Despite their progress, these methods rely on additional explicit 3D inputs. 
However, obtaining high-quality 3D data requires specialized sensors that are expensive and difficult to deploy, while in most real-world scenarios only 2D data (\textit{i.e.}, images or videos) is available. 

\noindent \textbf{3D Vision-Language Models with Only 2D Inputs.}
To overcome the dependency on explicit 3D data, recent works \cite{fan2025vlm,vgllm,wu2025spatial,zhang2026generalization,jiang2026towards} have shifted towards understanding 3D scenes solely from RGB videos. 
These methods typically employ 3D geometry encoders \cite{vggt,anysplat}, which are pre-trained on large-scale 2D-RGB and 3D paired data, to extract implicit 3D representations to obtain 3D geometric priors and inject them into VLMs to enhance their 3D capabilities. 
These implicit representations can encode the global 3D structural relationships of scenes, such as the overall arrangement of furniture in a room. 
In this work, we argue that this paradigm does not fully exploit the potential of 3D geometry encoders to enhance VLMs, as such implicit representations often encode coarse global spatial priors, which struggle with fine-grained 3D understanding tasks. 
In contrast, our VLM-IE3D introduces a new paradigm that incorporates global 3D spatial priors and detailed 3D structural information into VLMs by jointly leveraging implicit and explicit 3D geometric representations. 

\begin{figure*}[t]
\centering
\includegraphics[width=1.00\linewidth]{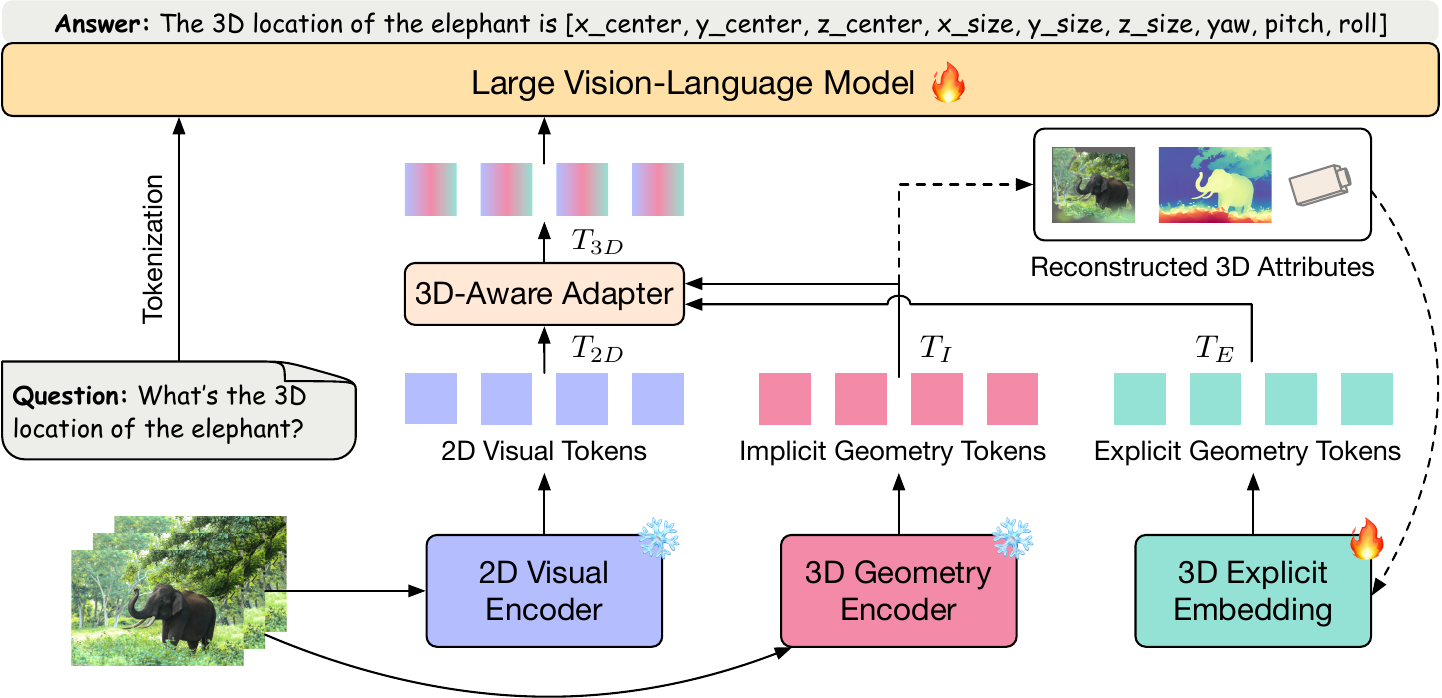}
\caption
{
Overview of the VLM-IE3D framework. 
VLM-IE3D directly processes RGB frames without explicit 3D inputs, using three streams: 
(1) a 2D visual encoder that extracts 2D visual tokens; 
(2) a 3D geometry encoder that generates Implicit Geometry Tokens (IGTs) for high-level 3D information; 
(3) a 3D explicit embedding that converts the reconstructed 3D explicit representations (\textit{e.g.}, depth maps and point clouds) into Explicit Geometry Tokens (EGTs) encoding fine-grained structural details. 
Finally, a 3D-aware adapter fuses these complementary tokens into unified and geometry-rich features for VLMs. }
\label{fig:pipeline}
\VspaceL
\end{figure*}

\section{Method}

The overview of our proposed VLM-IE3D is illustrated in Figure \ref{fig:pipeline}. 
Given an RGB video sequence $\left\{I^i\right\}_{i=1}^f$ and a natural language query $Q$, we first extract 2D visual tokens ${T}_{2D}\in \mathbb{R}^{f \times n \times c}$ using a 2D visual encoder. 
Here, $I^i\in \mathbb{R}^{h\times w\times 3}$, $f$ is the sequence length, $n=\left\lfloor\frac{h}{p}\right\rfloor \times \left\lfloor\frac{w}{p}\right\rfloor$, $p$ is the patch size, and $c$ is the channel dimension. 
The video sequence is also processed by a 3D geometry encoder \cite{anysplat} to produce Implicit Geometry Tokens (IGTs) $T_I\in \mathbb{R}^{f \times n\times c}$ along with corresponding 3D attributes.  
The reconstructed 3D attributes are further transformed by a 3D explicit embedding module to generate Explicit Geometry Tokens (EGTs) $T_E\in \mathbb{R}^{f \times n\times c}$. 
Subsequently, the three types of tokens are fused via a 3D-aware adapter, yielding enhanced 3D-aware visual tokens $T_{3D}$.
Finally, the pre-trained VLM backbone \cite{Qwen2.5-VL} takes $T_{3D}$ and $Q$ as input to generate the final response. 

\subsection{Implicit and Explicit Geometry Tokens}

We observe that the existing spatial VLMs~\cite{vgllm,fan2025vlm,wu2025spatial,huang2025mllms} rely solely on implicit 3D representations extracted by a 3D geometry encoder (see Figure \ref{fig:teaser} (a)), while overlooking the potential of explicit 3D representations. 
To address this limitation, we introduce Implicit Geometry Tokens (IGTs) and Explicit Geometry Tokens (EGTs). 
These two types of representations are inherently complementary: IGTs provide high-level and global implicit 3D spatial priors, while EGTs offer specific and fine-grained explicit 3D structural details, enabling more comprehensive 3D understanding and reasoning. 
In the following, we give details about the proposed IGTs and EGTs. 

\noindent \textbf{Implicit Geometry Tokens.}
3D geometry encoders~\cite{wang2024dust3r,tang2024mvdust3r, vggt,anysplat}, pre-trained on a diverse set of 3D geometric tasks (\textit{e.g.}, depth estimation and point reconstruction), have demonstrated strong capabilities in learning and representing 3D geometric information from 2D images. 
These models encode scenes into token sequences to enable 3D reconstruction and typically comprise three key components:
(1) an image encoder that extracts per-frame features, 
(2) a fusion decoder that employs alternating frame-wise and global self-attention mechanisms to facilitate cross-frame interaction, 
and (3) task-specific prediction heads for estimating 3D attributes such as depth maps. 

Inspired by these advances, we employ AnySplat \cite{anysplat} as our 3D geometry encoder and utilize the implicit 3D representations from the outputs of its fusion decoder as our IGTs $T_I\in \mathbb{R}^{f \times n \times c}$. 
IGTs encode high-level 3D spatial priors with a global understanding scope of scenes in the form of implicit latent vectors, such as the overall layout of a scene (\textit{e.g.}, the typical structure of a bedroom) and the spatial relationships between objects (\textit{e.g.}, ``a TV is mounted on the wall, and a sofa is placed in front of the TV''). 
By learning such representations, our method achieves strong generalization, enabling it to quickly adapt to 3D structural modeling of different scenes without explicit 3D inputs.

\noindent \textbf{Explicit Geometry Tokens.}
Despite the strengths of IGTs in learning global scene representations, their compressed and latent form makes it challenging for language models to interpret quantitative geometric properties, thereby limiting their effectiveness in quantitative spatial understanding and reasoning. 
To address this, we introduce Explicit Geometry Tokens (EGTs) that encode detailed 3D structural information with clear geometric interpretability from reconstructed explicit 3D attributes. 

Specifically, we obtain depth maps, camera poses, and 3D Gaussian splats directly from the corresponding prediction heads of the 3D geometry encoder \cite{anysplat}. 
The point maps are derived by back-projecting reconstructed depth maps into 3D coordinates using reconstructed camera poses. 
The well-defined 3D attributes $X_E\in \mathbb{R}^{f \times h \times w \times c^\prime}$ are then embedded into EGTs $T_E\in \mathbb{R}^{f \times n\times c}$ through a lightweight 3D explicit embedding module, where $c^\prime$ represents the channel dimension, which varies across different 3D attribute types. 
For efficiency and scalability, we avoid using heavy and complex 3D backbones (such as DepthAnything V2 \cite{yang2024depth}) and instead use a simple one-layer patch embedding followed by a two-layer MLP with minimal structural changes and training cost. 
EGTs thus serve as latent representations of explicit 3D structural data with spatial quantitative measurements, providing explicit geometric priors that complement the implicit priors captured by IGTs. 
These explicit priors focus on fine-grained 3D structural details and can provide VLMs with specific position information for better quantitative understanding and reasoning tasks. 

\begin{figure}[t]
\centering
\includegraphics[width=0.80\linewidth]{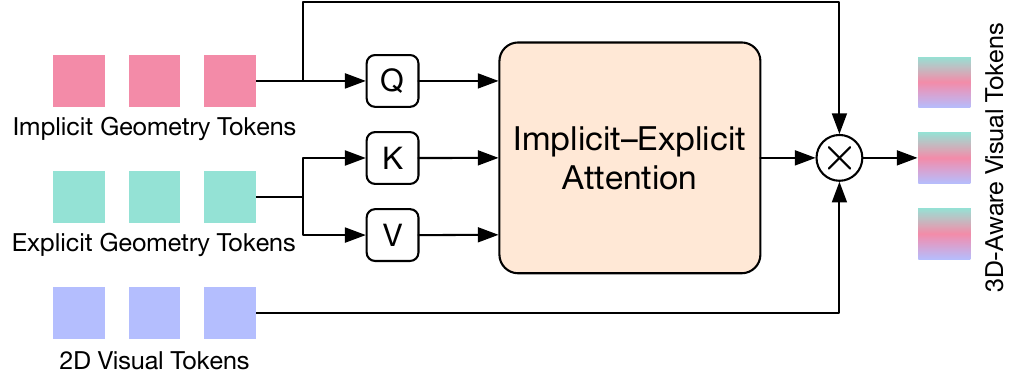}
\caption
{
Illustration of our 3D-aware adapter architecture. 
It takes the implicit and explicit 3D geometry tokens along with 2D visual tokens as input to generate the final 3D-aware visual tokens. 
The token compression operation is omitted for clarity. 
}
\VspaceL
\label{fig:adapter}
\end{figure}

\subsection{3D-Aware Adapter}

After the above process, we obtain three types of tokens: 2D visual tokens $T_{2D}$, IGTs $T_I$, and EGTs $T_E$. 
We then design a 3D-aware adapter to effectively integrate these three types of tokens, as shown in Figure \ref{fig:adapter}. 
Specifically, following the spatial merging strategy of Qwen2.5-VL \cite{Qwen2.5-VL}, we concatenate spatially adjacent $2\times 2$ features within each type of tokens and feed them into a two-layer MLP, producing compressed tokens $\{\tilde{T}_{2D}, \tilde{T}_{I}, \tilde{T}_{E}\}\in \mathbb{R}^{f \times m \times c}$, where $m{=}\left\lfloor\frac{h}{2p}\right\rfloor \times \left\lfloor\frac{w}{2p}\right\rfloor$. 

To bridge the gap between implicit and explicit 3D geometric representations, we develop an implicit–explicit attention (IEA) module to align and fuse these complementary representations. 
It performs interaction between IGTs and EGTs through a multi-head cross-attention (MCA) layer. 
Formally, the dot-product attention \cite{transformer} in MCA is expressed as:
\begin{equation}
    \operatorname{Attention}(Q, K, V)=\operatorname{Softmax}\left(Q K^T / \sqrt{d}\right) V,
\end{equation}
where queries $Q \in \mathbb{R}^{n_q \times d}$, keys $K \in \mathbb{R}^{n_k \times d}$, and values $V \in \mathbb{R}^{n_v \times d}$, with $d$ denoting the feature dimension.

For each frame $i$, MCA takes the compressed IGTs $\tilde{T}_I^{i}$ as queries and the compressed EGTs $\tilde{T}_E^{i}$ as keys and values, followed by a residual connection:
\begin{equation}
\tilde{T}_{3D}^{i} = \tilde{T}_I^{i} + \operatorname{MCA}(\tilde{T}_I^{i}, \tilde{T}_E^{i}, \tilde{T}_E^{i}),
\end{equation}
where $\operatorname{MCA}(\cdot)$ denotes the multi-head cross-attention function, and its inputs are queries, keys, and values. 

Next, we fuse 3D tokens $\tilde{T}_{3D}$ with compressed 2D visual tokens $\tilde{T}_{2D}$ via element-wise addition:
\begin{equation}
T_{3D} = \tilde{T}_{2D} + \tilde{T}_{3D}.
\label{equ:fusion}
\end{equation}

Through the 3D-aware adapter, our VLM-IE3D generates a unified 3D-aware visual representation $T_{3D}$, which is input into the pre-trained VLM together with text embeddings for final prediction. 
This design enables the model to jointly leverage both implicit and explicit 3D spatial relationships while preserving 2D visual semantics. 

\section{Experiments}

\subsection{Implementation Details}

We adopt Qwen2.5-VL-3B \cite{Qwen2.5-VL} as our VLM backbone and AnySplat \cite{anysplat} as our 3D geometry encoder. 
The 2D visual encoder is the same as that in Qwen2.5-VL \cite{Qwen2.5-VL}. 
We train the model for one epoch using the Adam optimizer with a warmup ratio of 0.03. 
The learning rate is gradually increased to 1e-5 and subsequently decayed to zero. 
The batch size is set to 1 per GPU. 
During training, the 2D visual encoder and 3D geometry encoder are frozen, while the VLM backbone and 3D explicit embedding remain trainable. 
All experiments are conducted on 8 H100 80G GPUs.  

The input image is rescaled and cropped to $392 \times 518$. 
We set the patch size $p=14$, the channel dimension $c=2048$, and the maximum sequence length $f=32$. 
The token number is $n=1036$, and the compressed token number is $m=252$. 
For explicit 3D representations, we explore three types of 3D attributes: depth maps ($c^\prime=1$), point maps ($c^\prime=3$), and 3D Gaussian splats ($c^\prime=86$). 
Since our method does not require ground truth 3D scene information, we follow common 3D geometry encoders \cite{vggt,anysplat} to adopt the first frame as the reference coordinate system for all tasks, except for 3D visual grounding where bounding boxes are represented in each frame's local coordinates. 
We use the depth map as the explicit 3D attribute in our EGTs, since it can be more readily obtained from RGB-D sensors for large-scale model training. 

\subsection{Comparison with State-of-the-Art Methods}

To verify the effectiveness of VLM-IE3D, we conduct extensive experiments on 3D scene understanding and spatial reasoning tasks to comprehensively evaluate its 3D understanding and reasoning capabilities.
For fair comparison, individual models are trained separately for the two types of tasks. 

\noindent \textbf{Results on 3D Scene Understanding.}
We train our model using multi-task learning on combined datasets and evaluate it on three fundamental 3D scene understanding tasks:

\noindent \textit{\textbf{(i)} 3D dense captioning} involves detecting 3D object proposals and generating descriptions based on object coordinates. 
We evaluate on the Scan2Cap \cite{scan2cap} benchmark. 
Following previous works \cite{vgllm,llava3d,video3dllm}, we utilize Mask3D-detected object proposals extracted by LEO \cite{leo}, and the model is then tasked with generating captions conditioned on object center coordinates. 

\noindent \textit{\textbf{(ii)} 3D visual grounding} aims to locate the first frame where a target object appears and its 3D bounding box in camera coordinates. 
We use the ScanRefer \cite{scanrefer} dataset, which comprises 36,665 object descriptions paired with axis-aligned bounding boxes across 562 indoor scans. 
Following \cite{vgllm,spar}, we refine predictions by matching the predicted bounding box with pre-detected object proposals. 

\noindent \textit{\textbf{(iii)} 3D video detection} predicts 3D bounding boxes for all visible objects across a sequence of consecutive frames. 
We use the dataset curated by \cite{vgllm} from EmbodiedScan \cite{embodiedscan}, which contains sequences of consecutive frames with corresponding object annotations in indoor scenes. 
Each sample consists of four consecutive frames captured at a rate of 1 FPS. 
Following \cite{vgllm}, we partition the data into 958 training and 243 evaluation scenes, randomly selecting 150 and 10 samples per scene for training and evaluation, respectively. 
We evaluate performance on 20 common object categories from daily life and report average precision, recall, and F1 score at an IoU threshold of 0.25, denoted as P$_{25}$, R$_{25}$, and F1$_{25}$, respectively. 

\begin{table}[t]
\scriptsize
\centering
\caption
{
Quantitative results on Scan2Cap. 
}
\VspaceS
\setlength{\tabcolsep}{3.05mm} 
\begin{tabular}{lc|cc}
\toprule [1pt]
\textbf{Method} & \makecell[c]{\textbf{3D Scene Input}} & \textbf{C@0.5$\uparrow$} & \textbf{M@0.5$\uparrow$} \\
\midrule [0.5pt]  

Scan2Cap \cite{scan2cap} &\cmark & 39.1 &22.0 \\
3DJCG \cite{3djcg} &\cmark & 49.5 &24.2  \\
D3Net \cite{d3net} &\cmark & 62.6 &25.7 \\

Vote2Cap-DETR \cite{vote2cap-detr} &\cmark  & 61.8 &26.2 \\
LL3DA \cite{ll3da}&\cmark & 65.2  & 26.0   \\
Grounded 3D-LLM \cite{grounded-3dllm}&\cmark & 70.2 &- \\
LEO \cite{leo} &\cmark   &72.4 &27.9 \\ 
Chat-Scene \cite{huang2024chatscene} &\cmark  &77.1 &- \\
LLaVA-3D \cite{llava3d} &\cmark  &79.2 & \best{30.2} \\
Video-3D LLM \cite{video3dllm} &\cmark & \best{80.0} &28.5 \\

\midrule [0.5pt]

Qwen2.5-VL-3B \cite{Qwen2.5-VL} &\xmark &58.0 &26.9 \\

VG~LLM \cite{vgllm} &\xmark & {78.6} & {28.6} \\

VLM-IE3D (Ours) &\xmark &\best{80.4} &\best{28.8} \\

\bottomrule [1pt]
\end{tabular}
\label{table:scan2cap}
\VspaceL
\end{table}

\begin{table}[t]
\scriptsize
\centering
\caption
{
Quantitative results on ScanRefer. 
The content in ``()'' indicates results with proposal refinement.
}
\VspaceS
\setlength{\tabcolsep}{2.60mm} 
\begin{tabular}{lc|cc}
\toprule [1pt]
\textbf{Method} & \makecell[c]{\textbf{3D Scene Input}} & \textbf{Acc@0.25} & \textbf{Acc@0.50} \\
\midrule [0.5pt]  

ScanRefer \cite{scanrefer} &\cmark & 37.3 & 24.3 \\
MVT \cite{mvt} &\cmark  & 40.8 & 33.3 \\
ViL3DRel \cite{chen2022vil3drel} &\cmark  & 47.9 &  37.7 \\
Grounded 3D-LLM \cite{grounded-3dllm} & \cmark & 47.9 & 44.1 \\ 
Chat-Scene \cite{huang2024chatscene} &\cmark  & 55.5 & 50.2 \\
LLaVA-3D \cite{llava3d} &\cmark  & 54.1 & 42.4 \\
Video-3D LLM \cite{video3dllm} &\cmark  & \best{58.1} & \best{51.7} \\ 

\midrule [0.5pt]  

SPAR \cite{spar}&\xmark  & 31.9 (48.8) & 12.4 (43.1) \\
Qwen2.5-VL-3B \cite{Qwen2.5-VL}&\xmark & 34.0 (50.7)   & 10.6 (44.7) \\
VG~LLM \cite{vgllm} &\xmark  & 36.4 (53.5)  & 11.8 (47.5)  \\
VLM-IE3D (Ours) &\xmark & \best{43.2} (55.4) & \best{16.9} (48.9) \\

\bottomrule [1pt]
\end{tabular}
\label{table:scanrefer}
\VspaceL
\end{table}

\begin{table}[t]
\scriptsize
\centering
\caption{Quantitative results on 3D video detection. Frames per second (FPS) was computed on a single H100 GPU. }
\VspaceS
\setlength{\tabcolsep}{3.20mm} 
\begin{tabular}{l|cc|ccc}
\toprule [1pt]
\textbf{Method} &Param (B) &FPS &P$_{25}$ & R$_{25}$ & F1$_{25}$ \\
\midrule [0.5pt]  

Qwen2.5-VL-3B \cite{Qwen2.5-VL} &3.09 &14&32.1 &30.1 &30.9 \\
VG~LLM \cite{vgllm} &3.13 &7 & 41.7 & 35.7 & 38.2  \\

\midrule [0.5pt]  

VLM-IE3D (Ours) &3.23 &6  &\best{44.2} &\best{41.9} &\best{42.8} \\

\bottomrule [1pt]
\end{tabular}
\label{table:detection}
\VspaceL
\end{table}

Our VLM-IE3D demonstrates strong performance across all three 3D scene understanding tasks, consistently outperforming methods using only 2D visual input while achieving competitive results with 3D-scene-input approaches. 

For 3D Dense Captioning (Scan2Cap), as shown in Table~\ref{table:scan2cap}, VLM-IE3D achieves the best C@0.5 among all compared methods (80.4) and the best M@0.5 among 2D-visual-input methods (28.8). 
Compared to baselines, VLM-IE3D shows substantial improvements of 22.4 in C@0.5 over Qwen2.5-VL-3B \cite{Qwen2.5-VL}, and 1.8 in C@0.5 over VG~LLM \cite{vgllm}, demonstrating that our IGTs and EGTs effectively enhance 3D spatial understanding capabilities. 

Table~\ref{table:scanrefer} presents results on the 3D visual grounding task (ScanRefer). 
VLM-IE3D achieves 43.2\% and 16.9\% accuracy at the IoU thresholds of 0.25 and 0.50, respectively, outperforming all methods without 3D input. 
The improvements of 9.2 in Acc@0.25 over Qwen2.5-VL-3B \cite{Qwen2.5-VL} and 6.8 over VG~LLM \cite{vgllm} demonstrate the superior object localization ability of our method. 
With proposal refinement, our method achieves 55.4\% and 48.9\% accuracy, narrowing the gap with 3D-scene-input methods. 
This indicates that while explicit 3D inputs provide advantages for precise localization, our implicit and explicit geometry encoding approach offers a competitive alternative. 

\begin{table*}[t]
\scriptsize
\centering
\caption{
    Quantitative comparison with state-of-the-art methods on VSI-Bench. 
}
\VspaceS
\resizebox{\textwidth}{!}{
\begin{tabular}{@{}l|cccccccc|c@{}}
\toprule[1pt]

& & {Obj. Count} & {Abs. Dist.} & {Obj. Size} & {Room Size} & {Rel. Dist.} & {Rel. Dir.} & {Route Plan} & {Appr. Order} \\

\cmidrule(lr){3-6} \cmidrule(lr){7-10}

\textbf{Models} & \textbf{Avg.} & \multicolumn{4}{c}{\textbf{Numerical Answer}} & \multicolumn{4}{c}{\textbf{Multiple-Choice Answer}} \\
\midrule[0.5pt]

\multicolumn{10}{l}{\textit{\textbf{Proprietary Models (API)}}} \\
\hspace{2mm}GPT-4o & 34.0 & 46.2 & 5.3 & 43.8 & 38.2 & 37.0 & 41.3 & 31.5 & 28.5 \\
\hspace{2mm}Gemini-1.5-Flash & 42.1 & 49.8 & 30.8 & 53.5 & 54.4 & 37.7 & 41.0 & 31.5 & 37.8 \\
\hspace{2mm}Gemini-1.5-Pro & 45.4 & 56.2 & 30.9 & \best{64.1} & 43.6 & \best{51.3} & \best{46.3} & 36.0 & 34.6 \\

\midrule[0.5pt]

\multicolumn{10}{l}{\textit{\textbf{Open-source Models}}} \\
\hspace{2mm}InternVL2-8B & 34.6 & 23.1 & 28.7 & 48.2 & 39.8 & 36.7 & 30.7 & 29.9 & 39.6 \\
\hspace{2mm}InternVL2-40B & 36.0 & 34.9 & 26.9 & 46.5 & 31.8 & 42.1 & 32.2 & 34.0 & 39.6 \\
\hspace{2mm}Qwen2.5-VL-3B & 30.6 & 24.3 & 24.7 & 31.7 & 22.6 & 38.3 & 41.6 & 26.3 & 21.2 \\
\hspace{2mm}Qwen2.5-VL-72B & 37.0 & 25.1 & 29.3 & 54.5 & 38.8 & 38.2 & 37.0 & 34.0 & 28.9  \\
\hspace{2mm}LongVA-7B & 29.2 & 38.0 & 16.6 & 38.9 & 22.2 & 33.1 & 43.3 & 25.4 & 15.7 \\
\hspace{2mm}VILA-1.5-40B & 31.2 & 22.4 & 24.8 & 48.7 & 22.7 & 40.5 & 25.7 & 31.5 & 32.9 \\
\hspace{2mm}VideoLLaMA3-7B & 35.8 & 41.9 & 23.5 & 42.2 & 27.1 & 39.4 & - & 32.0 & 31.4 \\
\hspace{2mm}LLaVA-OneVision-72B & 40.2 & 43.5 & 23.9 & \secondbest{57.6} & 37.5 & 42.5 & 39.9 & 32.5 & \secondbest{44.6} \\
\hspace{2mm}LLaVA-NeXT-Video-72B & 40.9 & 48.9 & 22.8 & {57.4} & 35.3 & 42.4 & 36.7 & 35.0 & \best{48.6} \\

\midrule[0.5pt]

\multicolumn{10}{l}{\textit{\textbf{Spatial-Enhanced Models}}} \\
\hspace{2mm}SAT-LLaVA-Video-7B & - & - & - & - & 47.3 & 41.1 & 37.1 & 36.1 & 40.4 \\
\hspace{2mm}SPAR-8B & 41.1 & - & - & - & - & - & - & - & - \\
\hspace{2mm}RynnEC-7B & 45.8 & 58.5 & 25.4 & 54.9 & 42.7 & 44.2 & - & \best{38.7} & 30.5 \\

\hspace{2mm}VG~LLM-4B & \secondbest{47.3} & \secondbest{66.0} & \secondbest{37.8} & {55.2} & \best{59.2} & 44.6 & \secondbest{45.6} & 33.5 & 36.4 \\

\hspace{2mm}VLM-IE3D-4B (Ours) &\best{47.6} &\best{67.5} &\best{38.5} &{55.0} &\secondbest{58.7} &\secondbest{47.7} &45.1 &\secondbest{36.6} &31.9 \\

\bottomrule[1pt]
\end{tabular}
}
\label{tab:vsi_bench}
\VspaceL
\end{table*}

Table~\ref{table:detection} shows results on the 3D video detection task. 
VLM-IE3D achieves 44.2\% P$_{25}$, 41.9\% R$_{25}$, and 42.8\% F1$_{25}$, representing improvements of 12.1, 11.8, and 11.9 over the baseline Qwen2.5-VL-3B \cite{Qwen2.5-VL}, and 2.5, 6.2, and 4.6 over VG~LLM \cite{vgllm}. 
These consistent gains across all metrics demonstrate that our proposed IGTs and EGTs effectively enhance the model's ability to detect and localize objects in 3D video scenes. 

\noindent \textbf{Efficiency and Complexity Analysis.}
Our method remains lightweight and efficient. As shown in Table~\ref{table:detection}, VG~LLM requires 3.13B trainable parameters, whereas VLM-IE3D has 3.23B. Despite incorporating the 3D explicit embedding module, the 3D-aware adapter, and the 3D attribute reconstruction step for EGTs, VLM-IE3D introduces only 0.10B additional parameters (a marginal 3.2\% increase). Furthermore, compared to the implicit baseline VG~LLM, our inference speed drops only marginally by 1 FPS (from 7 FPS to 6 FPS on a single H100 GPU). This slight decrease in speed is a highly acceptable trade-off considering the consistent performance improvements (\textit{e.g.}, +4.6 in F1$_{25}$ for 3D video detection compared to VG~LLM). The lightweight nature of our explicit geometry embedding (which relies on a simple MLP rather than a heavy deep encoder) ensures that the computational overhead remains negligible, facilitating potential deployment in real-world scenarios. 

\begin{figure*}[t]
\centering
\includegraphics[width=1.00\linewidth]{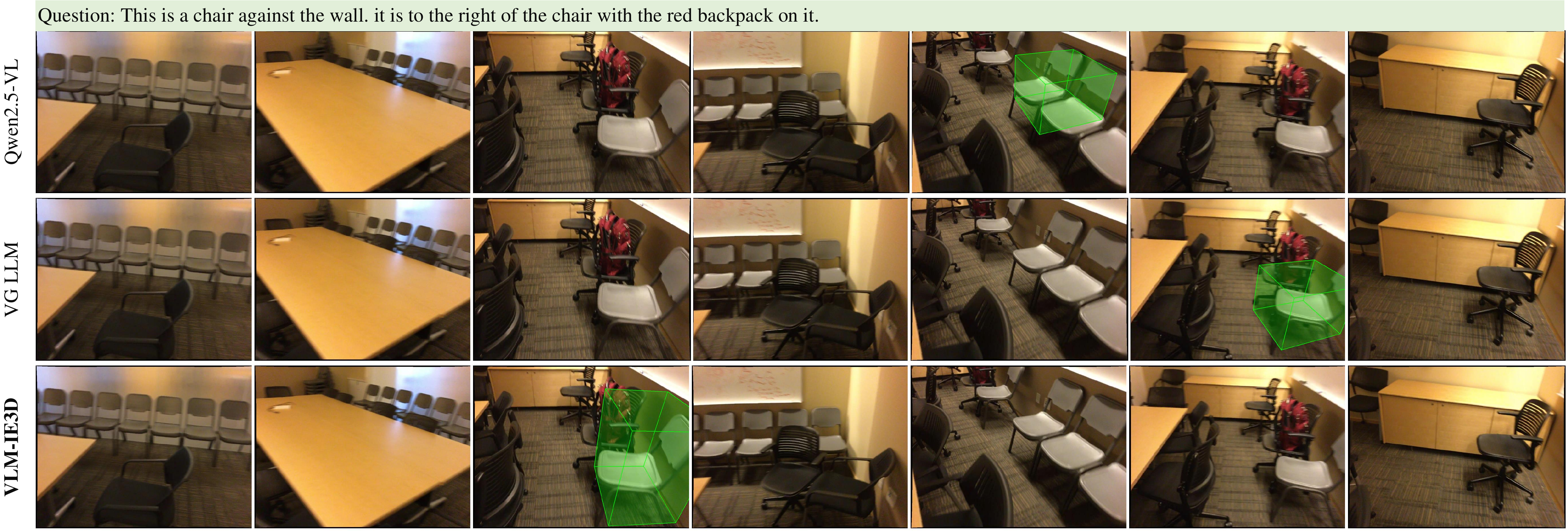}
\caption
{
Qualitative comparison of 3D visual grounding among Qwen2.5-VL, VG~LLM, and our VLM-IE3D. 
This task aims to localize the first frame in the video that contains the object described in the text. 
Our VLM-IE3D achieves superior localization accuracy and produces more precise object bounding boxes. 
}
\label{fig:vis_scanrefer}
\VspaceL
\end{figure*}

\begin{figure*}[t]
\centering
\includegraphics[width=1.00\linewidth]{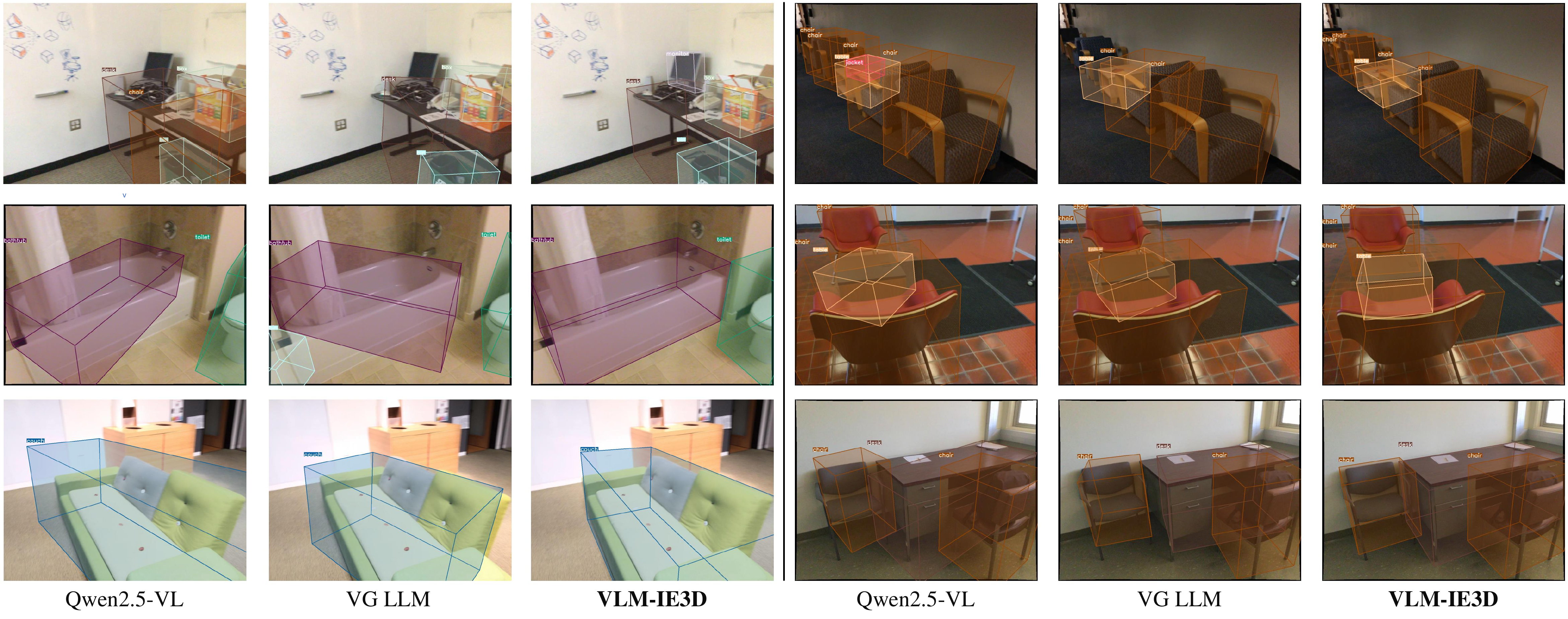}
\caption
{
Qualitative comparison of 3D video detection among Qwen2.5-VL, VG~LLM, and our VLM-IE3D. 
Our method produces more accurate 3D bounding box predictions with superior shape alignment and spatial localization, precisely capturing object geometry. 
}
\label{fig:vis}
\VspaceL
\end{figure*}

\noindent \textbf{Qualitative Results Analysis.}
Figure~\ref{fig:vis_scanrefer} and Figure~\ref{fig:vis} illustrate qualitative comparisons among Qwen2.5-VL \cite{Qwen2.5-VL}, VG~LLM \cite{vgllm}, and VLM-IE3D on 3D visual grounding and 3D video detection, respectively. As shown in Figure~\ref{fig:vis_scanrefer}, when given a complex natural language query referring to a target in a video, baseline models like Qwen2.5-VL and VG~LLM often struggle to accurately identify the exact first frame where the object appears, and they fail to precisely localize it in 3D space, producing bounding boxes that deviate from actual object boundaries. In contrast, VLM-IE3D correctly grounds the target in the temporal sequence and produces much tighter, more precise 3D bounding boxes. Similarly, in the 3D video detection task (Figure~\ref{fig:vis}), VLM-IE3D captures fine-grained local geometries better, significantly alleviating typical failure cases such as scale mismatch or floating predictions observed in the implicit-only baseline (VG~LLM). 
These qualitative results intuitively validate that complementing high-level IGTs with fine-grained structural EGTs effectively improves both the spatio-temporal and 3D geometric awareness of VLMs. 

\noindent \textbf{Results on Spatial Reasoning.}
We follow the training protocol of \cite{vgllm} for a fair comparison, utilizing subsets from SPAR-7M \cite{spar} (234K samples, 3\% of the full dataset) and the LLaVA-Hound split of LLaVA-Video-178K \cite{zhang2024video} (63K samples, 25\% of the original data). 
We then evaluate VLM-IE3D on VSI-Bench, a spatial reasoning benchmark that assesses egocentric-allocentric transformation and relational reasoning capabilities. 

Table~\ref{tab:vsi_bench} presents a comprehensive comparison of VLM-IE3D against state-of-the-art methods on VSI-Bench. 
Despite using only 4B parameters, our method achieves the best average performance of 47.6\%, outperforming all baselines including proprietary models like Gemini-1.5-Pro (45.4\%), and open-source models with significantly larger parameter counts such as LLaVA-NeXT-Video-72B (40.9\%). 
On numerical answer tasks, our model excels particularly in object counting (67.5\%), outperforming the previous best spatial-enhanced model VG~LLM by 1.5. 
For multiple-choice tasks, VLM-IE3D achieves competitive performance, particularly in tasks demanding high geometric perception where explicit structures are critical. 
For instance, on VSI-Bench (Relative Distance), we achieve a notable gain of +3.1 over VG~LLM (47.7\% vs. 44.6\%). 
These improvements demonstrate that explicit representations effectively capture fine-grained information to resolve complex spatial ambiguities where purely implicit baselines (VG~LLM) struggle, validating our motivation that combining implicit and explicit 3D representations provides complementary strengths for comprehensive spatial reasoning. 

\subsection{Ablation Study}

To verify the impact of each component and design in the proposed model, we conduct extensive ablation experiments on 3D video detection. 

\noindent \textbf{Implicit and Explicit Geometry Tokens.}
We adopt Qwen2.5-VL-3B \cite{Qwen2.5-VL}, which does not incorporate any 3D geometric information, as our baseline model and conduct ablation studies to investigate the effects of the proposed IGTs and EGTs. 
The results are summarized in Table~\ref{table:ablation}. 
The results demonstrate that incorporating either implicit or explicit 3D representations substantially improves VLM performance.
Specifically, equipping the baseline with EGTs improves F1$_{25}$ from 30.9 to 34.7, while incorporating IGTs yields a more significant improvement from 30.9 to 40.5.
These gains can be attributed to the high-level 3D priors captured by IGTs and the detailed structural information provided by EGTs, validating our core insight.
Notably, IGTs deliver larger improvements than EGTs, which is expected given that IGTs encode richer and more general 3D priors, offering more comprehensive geometric information.
When both implicit and explicit 3D representations are integrated, our method achieves further performance gains, reaching an F1$_{25}$ of 42.8. 
This represents an improvement of 11.9 over the baseline (from 30.9 to 42.8), 8.1 over the baseline with EGTs alone (from 34.7 to 42.8), and 2.3 over the baseline with IGTs alone (from 40.5 to 42.8). 
These results demonstrate that implicit and explicit 3D representations are complementary, and their joint integration effectively enhances the 3D capabilities of VLMs, validating our motivation. 

\begin{table}[t]
\scriptsize
\centering
\caption{Ablation study on implicit and explicit 3D feature representations. 
}
\VspaceS
\setlength{\tabcolsep}{6.20mm} 
\begin{tabular}{l|ccc}
\toprule [1pt]
\textbf{Method} & P$_{25}$ & R$_{25}$ & F1$_{25}$ \\
\midrule [0.5pt]  

Baseline \cite{Qwen2.5-VL} &32.1 &30.1 &30.9 \\ 

\midrule

+ EGTs &36.1 &33.7 &34.7 \\
+ IGTs &41.8 &39.7 &40.5 \\
+ IGTs + EGTs &{44.2} &{41.9} &{42.8} \\

\bottomrule [1pt]
\end{tabular}
\label{table:ablation}
\VspaceL
\end{table}

\begin{table}[t]
\scriptsize
\centering
\caption
{
Ablation study on different fusion strategies for IGTs and EGTs. 
}
\VspaceS
\setlength{\tabcolsep}{4.10mm} 
\begin{tabular}{l|ccc}
\toprule [1pt]
\textbf{Method} & P$_{25}$ & R$_{25}$ & F1$_{25}$ \\
\midrule [0.5pt]

VLM-IE3D, Concat    &42.7 &40.8 &41.5 \\
VLM-IE3D, Addition  &43.4 &41.9 &42.4 \\
VLM-IE3D, Weighted  &42.3 &40.4 &41.2 \\
VLM-IE3D, Proposed IEA &{44.2} &{41.9} &{42.8} \\

\bottomrule [1pt]
\end{tabular}
\label{table:fusion}
\VspaceL
\end{table}

\noindent \textbf{Fusion Strategy.}
To investigate the effect of different fusion mechanisms between implicit and explicit representations (\textit{i.e.}, IGTs and EGTs), we conduct an ablation study as shown in Table~\ref{table:fusion}. 
We ablate the following fusion strategies:
{\textbf{(i)} Concat:} concatenation of features followed by a linear projection to align their dimensions. 
{\textbf{(ii)} Addition:} element-wise addition of the two feature maps. 
{\textbf{(iii)} Weighted:} element-wise addition with two learnable weights.
{\textbf{(iv)} Proposed IEA:} our proposed IEA module, which dynamically exchanges information between IGTs and EGTs. 
Among these strategies, the proposed IEA yields the best performance, indicating its effectiveness in integrating implicit and explicit representations.
We also explore different fusion strategies for 2D and 3D token fusion as described in Eq.~\ref{equ:fusion}. 
Empirical results show that the direct and non-parametric addition operation is sufficient to achieve strong performance. 
Therefore, we choose the proposed IEA for fusing IGTs and EGTs and the addition operation for fusing 2D and 3D tokens. 

\noindent \textbf{Explicit 3D Attribute.}
We further investigate the impact of different explicit 3D attributes, including depth maps, point maps, and 3D Gaussians. 
As shown in Table~\ref{table:attributes}, incorporating any of these explicit 3D representations consistently improves performance across all metrics, achieving highly comparable results (F1$_{25}$ scores ranging from 42.5\% to 42.8\%). 
This indicates that as long as the explicit representation provides physically meaningful spatial measurements, it can effectively compensate for the lack of fine-grained structures in IGTs. 
Consequently, we opt for depth maps in our default setting, as they are the most accessible and parameter-efficient representation among the three. These results validate that different explicit 3D representations can serve as robust geometric complements to IGTs, thereby consistently enhancing the spatial awareness of VLMs.

\begin{table}[t]
\scriptsize
\centering
\caption{Ablation study on different explicit 3D attributes. 
}
\VspaceS
\setlength{\tabcolsep}{4.60mm} 
\begin{tabular}{l|ccc}
\toprule [1pt]
\textbf{Method} & P$_{25}$ & R$_{25}$ & F1$_{25}$ \\
\midrule [0.5pt]

Baseline \cite{Qwen2.5-VL} &32.1 &30.1 &30.9 \\

\midrule

+ IGTs &41.8 &39.7 &40.5 \\
+ IGTs + EGTs (Point) &44.3 &41.7 &42.6 \\
+ IGTs + EGTs (Depth) &{44.2} &{41.9} &{42.8} \\
+ IGTs + EGTs (Gaussian) &43.9 &41.6 &42.5 \\

\bottomrule [1pt]
\end{tabular}
\label{table:attributes}
\VspaceL
\end{table}

\begin{table}[t]
\scriptsize
\centering
\caption
{
Ablation study on different design choices of the 3D explicit embedding. 
}
\VspaceS
\setlength{\tabcolsep}{3.40mm} 
\begin{tabular}{l|ccc}
\toprule [1pt]
\textbf{Method} & P$_{25}$ & R$_{25}$ & F1$_{25}$ \\
\midrule [0.5pt]  

VLM-IE3D, None &41.8 &39.7 &40.5 \\
VLM-IE3D, Pooling &43.6 &41.0 &42.0 \\
VLM-IE3D, Positional Embedding &44.0 &41.5 &42.5 \\
VLM-IE3D, Deep Encoder  &38.5 &34.0 &35.9 \\

\midrule [0.5pt]  

VLM-IE3D &{44.2} &{41.9} &{42.8} \\

\bottomrule [1pt]
\end{tabular}
\label{table:embedding}
\VspaceL
\end{table}

\noindent \textbf{3D Explicit Embedding.}
Table~\ref{table:embedding} presents an ablation study examining different design choices for the 3D explicit embedding module in VLM-IE3D.
We evaluate four alternative approaches for embedding explicit geometry:
{\textbf{(i)} None:} removing the 3D explicit embedding and EGTs entirely.
{\textbf{(ii)} Pooling:} replacing the one-layer patch embedding in our method with the average pooling operation. 
{\textbf{(iii)} Positional Encoding:} using average pooling with sinusoidal positional embedding \cite{transformer}. 
{\textbf{(iv)} Deep Encoder:} treating depth maps as images and utilizing DepthAnything V2 \cite{yang2024depth} as a deep encoder. 
The results demonstrate the effectiveness of our simple and lightweight design for the 3D explicit embedding module, which introduces only 0.008B parameters (a negligible 0.25\% of the 3.23B trainable parameters) while achieving significant performance gains. 
Surprisingly, we observe that a deep encoder architecture is challenging to train and yields suboptimal performance. 
We hypothesize that complex deep models tend to over-process the inputs and extract high-level abstract semantic features. 
This essentially overlaps with the role of IGTs, causing feature redundancy. In contrast, explicit geometry requires a direct mapping of spatial coordinates. 
Our lightweight design accurately preserves these fine-grained details without over-abstraction, serving as an ideal complement to the abstraction-heavy IGTs. 

\noindent \textbf{3D Geometry Encoder.}
To evaluate the generalization capability of our method across different 3D geometry encoders, we conduct ablation studies using VGGT \cite{vggt}, AnySplat \cite{anysplat}, and $\pi^3$ \cite{wang2025pi}. 
As shown in Table~\ref{table:encoders}, all tested encoders consistently improve performance over the baseline model \cite{Qwen2.5-VL}. 
We also replace the depth maps from AnySplat with those reconstructed by DepthAnything V2 \cite{yang2024depth}, yielding IGTs and EGTs from different source models. 
Notably, this variant also exhibits strong performance, demonstrating that IGTs and EGTs can be extracted from separate models.  
For efficiency and simplicity, we use a single 3D geometry encoder to obtain both IGTs and EGTs in our final framework. 
These results demonstrate the robust generalization capability of our VLM-IE3D framework and highlight its flexibility for seamless integration with diverse 3D geometry encoders, suggesting strong potential for leveraging future advancements of 3D geometry encoders. 

\begin{table}[t]
\scriptsize
\centering
\caption
{
Ablation study on different 3D geometry encoders. 
}
\VspaceS
\setlength{\tabcolsep}{2.00mm} 
\begin{tabular}{l|ccc}
\toprule [1pt]
\textbf{Method} & P$_{25}$ & R$_{25}$ & F1$_{25}$ \\
\midrule [0.5pt]  

Baseline \cite{Qwen2.5-VL} &32.1 &30.1 &30.9 \\

\midrule

VLM-IE3D, $\pi^3$  &43.6 &41.2 &42.1 \\
VLM-IE3D, VGGT &43.1 &40.9 &41.7 \\
VLM-IE3D, AnySplat  &{44.2} &{41.9} &{42.8} \\
VLM-IE3D, AnySplat w. DepthAnything V2 &43.8 &41.5 &42.5 \\

\bottomrule [1pt]
\end{tabular}
\label{table:encoders}
\VspaceL
\end{table}

\section{Conclusion}

This paper presents VLM-IE3D, a unified framework that enhances the 3D spatial awareness of VLMs by integrating both implicit and explicit 3D geometric representations from purely 2D visual inputs. 
Through the complementary design of IGTs and EGTs and the effective fusion of the 3D-aware adapter, our method injects strong 3D inductive biases into VLMs and achieves both holistic and fine-grained 3D understanding and reasoning without requiring 3D inputs. 
Experiments on various 3D scene understanding and spatial reasoning tasks validate the effectiveness of the proposed VLM-IE3D, showing superior performance over 2D-visual-input methods and competitive results with 3D-scene-input methods. 

{\small
\noindent \textbf{Acknowledgements.}
This study is funded by the Ministry of Education Singapore, under the Tier-2 project scheme with project number MOET2EP20123-0003.
}

{
    \small
    \bibliographystyle{splncs04}
    \bibliography{ref}
}

\clearpage
\appendix
\gdef\theHsection{\appendixname.\thesection} 

{\noindent\Large\textbf{Supplementary Material}}
\newline

This supplementary material covers the following details:
\begin{itemize}
\item The prompts for 3D scene understanding tasks (Sec.~\ref{sec:Prompt}). 
\item Additional visualization results (Sec.~\ref{sec:visualization_results}). 
\end{itemize}

\section{Example Prompts}
\label{sec:Prompt}

The detailed prompts for the three 3D scene understanding tasks are provided in Table~\ref{tab:3d_prompt}. 

\section{Additional Visualization Results}
\label{sec:visualization_results}

\begin{figure*}[t]
\centering
\refstepcounter{table}
\caption*{\textbf{Table~\thetable.} The prompts for 3D scene understanding tasks.}
\label{tab:3d_prompt}
\begin{minipage}{\columnwidth}\centering
\begin{tcolorbox}[
    colback=cyan!2,
    colframe=cyan!40!black,
    boxrule=0.6pt,
    sharp corners,
    arc=1pt,
    left=6pt, right=6pt, top=5pt, bottom=5pt,
    width=\textwidth
]
\raggedright
{\fontsize{8.3pt}{9.8pt}\selectfont

\textcolor{cyan!60!black}{\textbf{3D Visual Grounding}} \\[2pt]

\textbf{Question:}  
Frame-0: \PredSty{\texttt{<image>}} Frame-1: \PredSty{\texttt{<image>}} Frame-2: \PredSty{\texttt{<image>}} Frame-3: \PredSty{\texttt{<image>$\cdots$}} \\
Localize the first clear frame in the video showing the object described in the text. \\
\textit{Text:} There is an upper cabinet in the room. It has a fridge to its right.\\[1mm]

\textbf{Answer:} \\
\textasciigrave\textasciigrave\textasciigrave JSON \\
\texttt{\{}"frame": 7, "bbox\_3d": [-0.2, -0.07, 2.3, 1.23, 0.33, 0.84, -0.5, 1.53, -2.1]\texttt{\}} 
\textasciigrave\textasciigrave\textasciigrave \\

\medskip
\hrule
\medskip

\textcolor{cyan!60!black}{\textbf{3D Dense Captioning}} \\[2pt]

\textbf{Question:}  \PredSty{\texttt{<image><image><image><image>$\cdots$}} \\
Carefully watch the video and describe the object located at [3.38, -0.19, 2.07] in detail. \\[1mm]
\textbf{Answer:} The object is a cabinet on the right side of the room at the end of the walkway with a red chair against the wall. It is white and has a swivel chair sitting in front of it. \\

\medskip
\hrule
\medskip

\textcolor{cyan!60!black}{\textbf{3D Video Object Detection}} \\[2pt]

\textbf{Question:}  \PredSty{\texttt{<image><image><image><image>$\cdots$}} \\
Detect the 3D bounding boxes in the camera coordinate system of the first frame.\\
Output a JSON list where each entry contains the object name in "label" and its 3D bounding box in "box\_3d". \\
The 3D bounding box format should be [x\_center, y\_center, z\_center, x\_size, y\_size, z\_size, yaw, pitch, roll]. \\[1mm]

\textbf{Answer:} \\
\textasciigrave\textasciigrave\textasciigrave JSON \\
\texttt{[} \\
$\qquad$ \texttt{\{}"label": "socket", "bbox\_3d": [0.85, -0.54, 0.5, 0.01, 0.11, 0.11, 1.34, 0.74, -3.11]\texttt{\}}, \\
$\qquad$ \texttt{\{}"label": "doorframe", "bbox\_3d": [0.3, -0.62, 0.86, 0.1, 0.9, 2.08, 1.35, 0.74, -3.11]\texttt{\}}, \\
$\qquad \cdots$ \\
\texttt{]} \textasciigrave\textasciigrave\textasciigrave
}
    
\end{tcolorbox}
\vspace{4mm}
\end{minipage}
\includegraphics[width=0.96\linewidth]{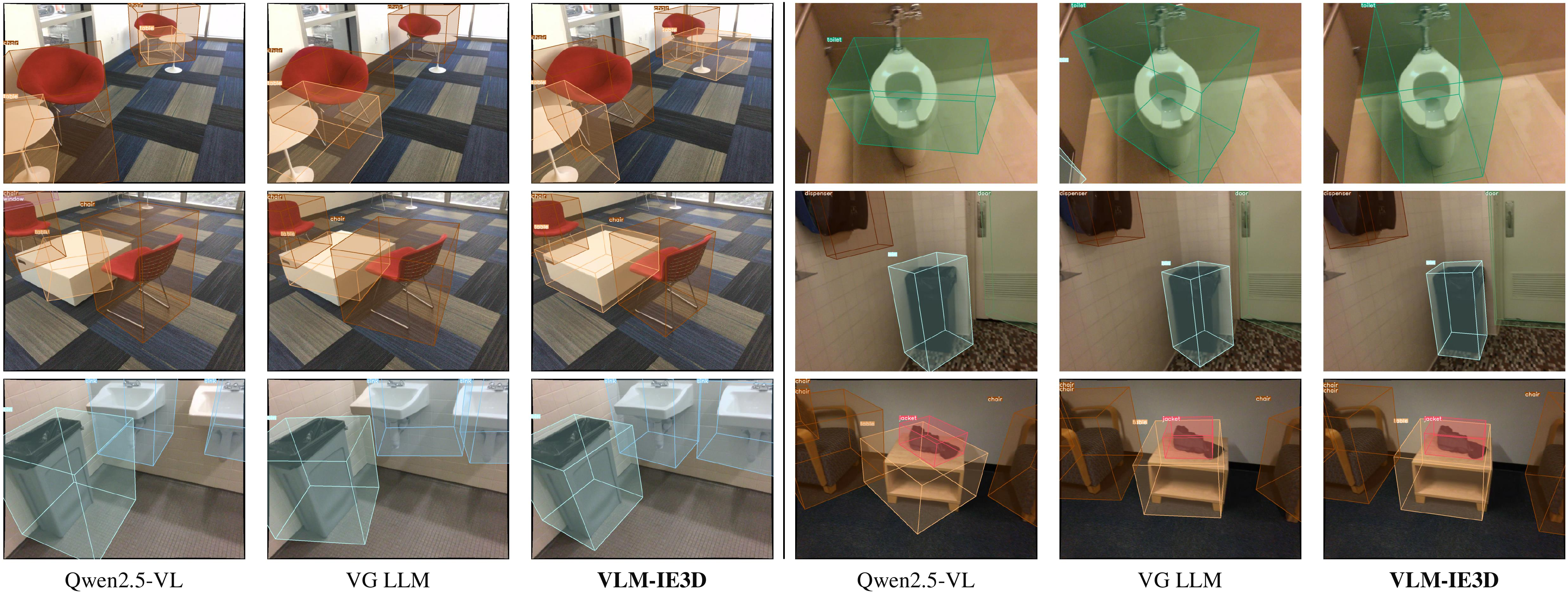}
\caption
{
Qualitative comparison of 3D video detection among Qwen2.5-VL, VG~LLM, and our VLM-IE3D. 
}
\label{fig:vis_supp}
\end{figure*}

\begin{figure*}[t]
\centering
\includegraphics[width=0.92\linewidth]{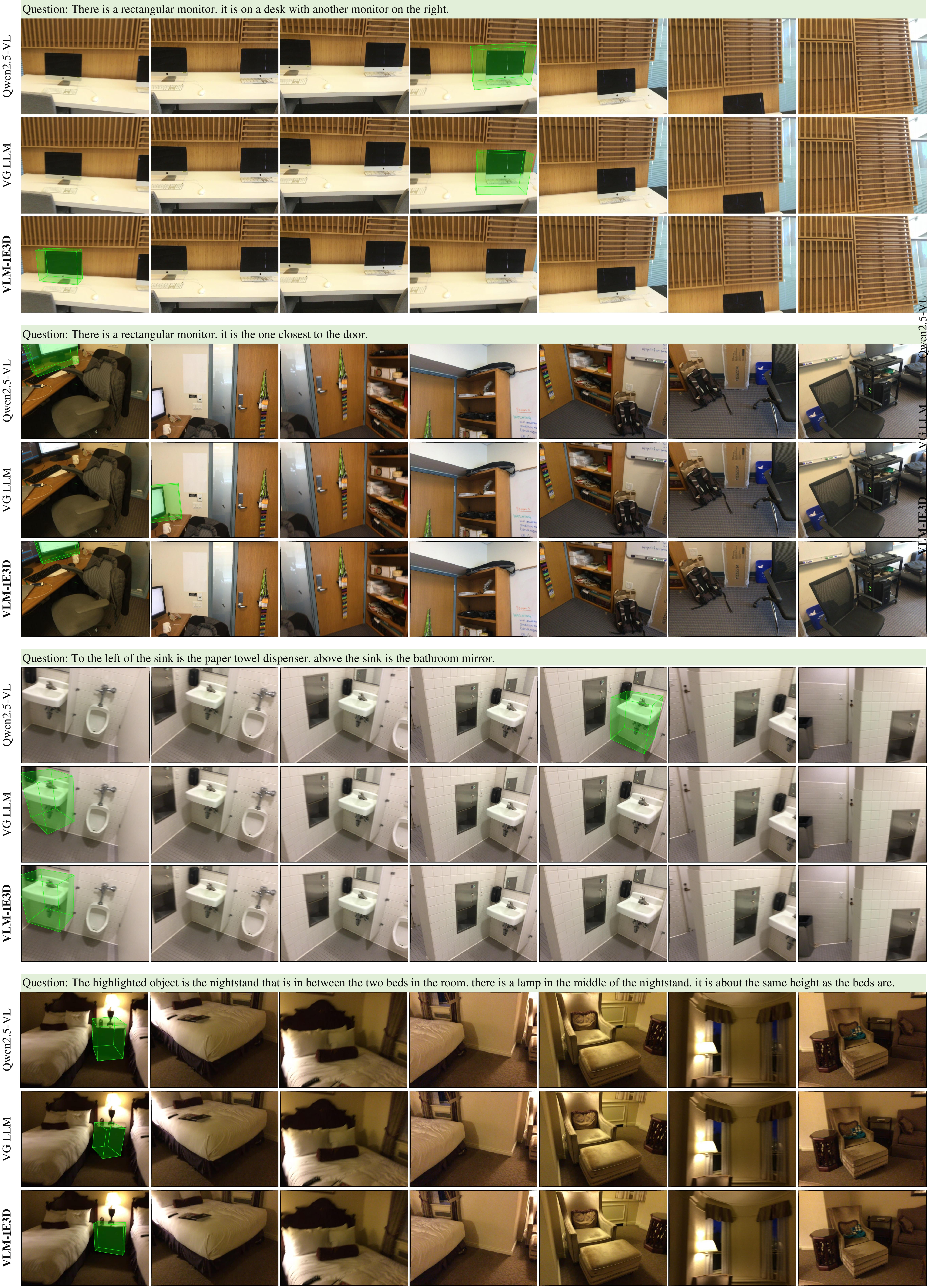}
\caption
{
Qualitative comparison of 3D visual grounding among Qwen2.5-VL, VG~LLM, and our VLM-IE3D. 
This task aims to localize the first frame in the video that contains the object described in the text. 
}
\label{fig:vis_scanrefer_supp}
\end{figure*}

We provide qualitative results in Figures~\ref{fig:vis_supp} and~\ref{fig:vis_scanrefer_supp}, comparing VLM-IE3D with Qwen2.5-VL \cite{Qwen2.5-VL} and VG~LLM \cite{vgllm} on 3D video detection and 3D visual grounding. 
VLM-IE3D achieves more accurate localization of objects given natural language queries and produces tighter bounding boxes with improved detection precision. 
These visual results confirm that incorporating implicit and explicit 3D geometric priors via IGTs and EGTs substantially enhances the 3D capabilities of VLMs, leading to more robust object grounding and detection performance. 

\end{document}